\documentclass[11pt]{article}
\usepackage[utf8]{inputenc}
\usepackage[T1]{fontenc}
\usepackage{amsmath,amssymb}
\usepackage{booktabs}
\usepackage{graphicx}
\usepackage{hyperref}
\usepackage{natbib}
\usepackage[margin=1in]{geometry}
\usepackage{microtype}

\title{Repetition Without Exclusivity:\\Scale Sensitivity of Referential Mechanisms\\in Child-Scale Language Models}
\author{JP Cacioli\\Independent Researcher\\Melbourne, Australia}
\date{}

\begin{document}
\maketitle

\begin{abstract}
We present the first systematic evaluation of mutual exclusivity (ME)---the bias to map novel words to novel referents---in text-only language models trained on child-directed speech. We operationalise ME as referential suppression: when a familiar object is relabelled in a two-referent discourse context, ME predicts decreased probability of the labelled noun at a subsequent completion position. Three pilot findings motivate a pre-registered scale-sensitivity experiment: (1) a masked language model (BabyBERTa) is entirely insensitive to multi-sentence referential context, precluding ME evaluation; (2) autoregressive models show robust repetition priming---the opposite of ME---when familiar nouns are re-labelled; and (3) a novel context-dependence diagnostic reveals that apparent ME-like patterns with nonce tokens are fully explained by embedding similarity, not referential disambiguation. In the confirmatory experiment, we train 45 GPT-2-architecture models (2.9M, 8.9M, and 33.5M parameters; 5, 10, and 20 epochs on AO-CHILDES; 5 seeds each) and evaluate on a pre-registered ME battery. Anti-ME repetition priming is significant in all 9 cells of the model grid (85--100\% of items; all $p < 2.4 \times 10^{-13}$). Priming attenuates with improved language modelling (Spearman $\rho = -0.533$, $p = 0.0002$) but never crosses zero across a 3.8$\times$ perplexity range. The context-dependence diagnostic replicates in all 9 cells, and dose--response priming increases with repetitions in 8/9 cells (all trend $p < 0.002$). These findings indicate that, under child-scale data and architectures, distributional learning on child-directed speech produces repetition-based reference tracking rather than lexical exclusivity: priming attenuates toward a non-zero asymptote but does not invert. We connect this dissociation to the grounded cognition literature, where ME does emerge in visually grounded models, and argue that referential grounding may be a necessary ingredient for ME---an empirical claim about required input structure, not a nativist one.
\end{abstract}

\section{Introduction}

The BabyLM paradigm trains language models on child-scale data and evaluates linguistic competence \citep{warstadt2023findings,charpentier2025findings}. Evaluations focus on syntax \citep[BLiMP;][]{warstadt2020blimp} and natural language understanding \citep[GLUE;][]{wang2018glue}. But referential pragmatics---how children use words to pick out objects in the world---remains untested. We address this gap using mutual exclusivity (ME) as the test case.

ME is the bias to map novel words to novel referents rather than objects that already have a name \citep{markman1988childrens}. Intuitively, if a child sees a familiar dog and an unfamiliar object and hears ``look at the dax!'', ME leads the child to assume ``dax'' refers to the unfamiliar object---because the dog already has a name. The bias is central to early word learning, developmentally graded and modulated by vocabulary size \citep{lewis2020role}, weaker in bilinguals \citep{byersheinlein2009monolingual}, and has been studied computationally in grounded models \citep{gulordava2020deep,ohmer2022mutual,nortje2024mutual,oneata2025mutual,thai2025mebench}. A key feature of all computational work showing positive ME results is that the model has referents---images, object representations, or symbolic referent sets---to which labels can be mapped.

BabyLM-class models are text-only. Can ME emerge from distributional learning on child-directed speech alone, without referential grounding? This is not a question about whether text-only models \emph{should} exhibit ME---it is a question about whether distributional statistics are \emph{sufficient} to induce it. We test this across 45 autoregressive models spanning a 12$\times$ parameter range and 3.8$\times$ perplexity range, plus a masked language model \citep[BabyBERTa;][]{huebner2021babyberta}. We find no evidence of ME at any scale tested; instead, models develop repetition priming that attenuates with improved language modelling but never reverses sign. We provide a mechanistic account grounded in the statistics of child-directed speech and introduce a context-dependence diagnostic that rules out artefactual ME from embedding similarity---a reusable methodological contribution for prompt-based pragmatic evaluation.

\section{Background}

\subsection{ME in Developmental Psychology}

ME was first characterised by \citet{markman1988childrens} as a constraint on word learning: children assume that each object has only one label, so a novel word is mapped to the unnamed object rather than one that already has a name. The bias emerges around 17 months \citep{halberda2003development}, strengthens with vocabulary growth \citep{lewis2020role}, and is modulated by language experience---bilingual children show weaker ME, consistent with their routine experience of multiple labels for the same referent \citep{byersheinlein2009monolingual}. An important distinction drawn by \citet{mcmurray2012word} separates referent selection (an in-the-moment disambiguation) from word learning (retention over time); our evaluation targets the former.

Bayesian accounts formalise ME as a consequence of probabilistic inference over word--referent mappings \citep{xu2007word}. Under this framework, the critical computational ingredient is the referent set: the learner must represent that there are distinct objects in the world, each requiring a label. The size principle---that a hypothesis consistent with fewer examples is more likely under strong sampling---naturally produces one-to-one label--referent mappings.

\subsection{ME in Computational Models}

Cross-situational neural models trained on symbolic and image data exhibit ME when lexical competition is built into the learning or selection process \citep{gulordava2020deep}. Pragmatic agents that combine Rational Speech Act reasoning with gradient-based learning also develop ME \citep{ohmer2022mutual}, as do meta-learning systems designed for few-shot word learning \citep{gandhi2020mutual}.

The most relevant recent work comes from visually grounded speech (VGS) models. \citet{nortje2024mutual} showed that ME emerges from contrastive learning on speech--image pairs, with stronger ME when models have more visual prior knowledge. Extending this to the bilingual case, \citet{oneata2025mutual} found that bilingual VGS models exhibit weaker ME---paralleling human developmental data. \citet{thai2025mebench} introduced MEBench and reported that even state-of-the-art vision-language models show only weak ME bias.

The key insight from this literature is that every positive ME result involves models with referents. ME has never been evaluated in text-only models trained on naturalistic child-directed speech.

\subsection{Child-Scale Language Models}

The BabyLM Challenge \citep{warstadt2023findings,charpentier2025findings} trains models on corpora of 10M--100M words---approximating the linguistic input available to children---and evaluates them on syntactic benchmarks (BLiMP, Zorro) and NLU tasks (GLUE). BabyBERTa \citep{huebner2021babyberta} trains a RoBERTa-base model on AO-CHILDES, a corpus of American-English child-directed speech (${\sim}$4M tokens). These evaluations focus on syntax and NLU; referential pragmatics is untested. We fill this gap.

\section{Method}

\subsection{Pre-Registration and Transparency}

We conducted a pilot study on a single autoregressive model and BabyBERTa (described in \S\ref{sec:pilot}) that yielded three mechanistic findings. Based on these exploratory results, we pre-registered a confirmatory scale-sensitivity experiment on the Open Science Framework (OSF; \url{https://osf.io/zu7af}) before training any additional models. The pre-registration specifies four hypotheses (\S\ref{sec:hypotheses}), statistical tests, significance thresholds, and the exact training grid. Train and evaluation scripts were archived in the registration snapshot and are publicly available at \url{https://github.com/synthiumjp/study2-me-eval}. Pilot findings are reported as exploratory with full transparency about the analysis path.

\subsection{Corpus}

AO-CHILDES \citep{huebner2021babyberta} contains 893,989 sentences (${\sim}$4.07M word tokens, 6.27M BPE tokens) of American-English child-directed speech from the CHILDES database \citep{macwhinney2000childes}. The corpus is overwhelmingly composed of single-referent naming contexts; multi-referent disambiguation constructions (e.g., ``the X or the Y'') are rare.

A property of CDS that is directly relevant to our findings is its high rate of noun repetition. Among the nine concrete nouns used in our evaluation battery (\emph{ball, book, car, cup, hat, dog, cat, fish, bird}), the probability of a noun reappearing within the next 3 sentences averages 36.4\% (range: 30.6\% for \emph{dog} to 42.2\% for \emph{ball}), and every test noun participates in hundreds to thousands of 10-sentence windows containing 3 or more mentions (e.g., \emph{ball}: 4,236 windows; \emph{cup}: 1,050). This repetition structure---characteristic of how caregivers scaffold naming (``that's a ball\,.\,see the ball?\,.\,can you get the ball?'')---provides a strong distributional signal favouring repetition priming. By contrast, sentences containing two or more of our target nouns account for only 524 of 893,989 sentences (0.06\%), and the majority of these are compound expressions (\emph{kitty cat}) rather than contrastive reference. The distributional environment of CDS thus strongly favours label repetition over label contrast.

\subsection{Models}

\paragraph{Masked LM (pilot only).} BabyBERTa \citep{huebner2021babyberta}: RoBERTa-base architecture, 8K vocabulary, trained on AO-CHILDES individual sentences. We use the published checkpoint.

\paragraph{Autoregressive models (confirmatory).} GPT-2 architecture \citep{radford2019language} trained from scratch on AO-CHILDES. We train a $3 \times 3$ grid of model sizes and training durations, with 5 random seeds per cell (45 models total):

\begin{table}[h]
\centering
\small
\begin{tabular}{lcccc}
\toprule
Size & Layers & Dim & Heads & Parameters \\
\midrule
Small & 4 & 128 & 4 & 2,862,848 \\
Medium & 6 & 256 & 8 & 8,878,080 \\
Large & 8 & 512 & 8 & 33,498,112 \\
\bottomrule
\end{tabular}
\end{table}

All models share a BPE vocabulary of 8,020 tokens (8,000 merges + 20 nonce tokens added with space-prefix encoding). Training durations: 5, 10, and 20 epochs. Optimizer: AdamW, learning rate $5 \times 10^{-4}$, cosine schedule, batch size 32, context length 128 tokens. Seeds 0--4. Total wall time: 31.4 hours on an AMD RX 7900 GRE (16\,GB) via DirectML.

Parameter counts differ from pre-registration estimates (${\sim}$2M, ${\sim}$7M, ${\sim}$30M) due to the embedding matrix dimension with the 8,020-token vocabulary. We report actual counts throughout.

\subsection{ME Evaluation Battery}

\subsubsection{Familiar--Familiar Suppression Track (H1)}

We test whether labelling one of two introduced referents as a familiar noun increases or decreases the probability of predicting that noun at a subsequent completion position.

\noindent\emph{Baseline:} ``there is a [FAM$_1$] and a [FAM$_2$]\,. this is a \underline{\hspace{1cm}}''\\
\emph{ME condition:} ``there is a [FAM$_1$] and a [FAM$_2$]\,. that is a [FAM$_1$]\,. this is a \underline{\hspace{1cm}}''

ME predicts $P(\text{FAM}_1 \mid \text{ME}) < P(\text{FAM}_1 \mid \text{baseline})$ (suppression) and $P(\text{FAM}_2 \mid \text{ME}) > P(\text{FAM}_2 \mid \text{baseline})$ (boost). Anti-ME predicts the opposite. We test 10 noun pairs in both directions (20 items per model seed). Log-probabilities are extracted via teacher-forced next-token prediction.

\subsubsection{Context-Dependence Diagnostic (H2)}

To distinguish genuine ME from embedding-level artefacts, we evaluate nonce-token predictions under five conditions:

\begin{itemize}
\setlength\itemsep{0pt}
\item \textbf{full\_context}: ``there is a [FAM] and a [NONCE$_1$]\,. the [NONCE$_2$] is the \underline{\hspace{0.8cm}}'' (standard ME design)
\item \textbf{swap\_context}: Same with FAM and NONCE$_1$ order reversed
\item \textbf{nonce\_only}: ``there is a [NONCE$_1$]\,. the [NONCE$_2$] is the \underline{\hspace{0.8cm}}'' (no familiar noun)
\item \textbf{fam\_only}: ``there is a [FAM]\,. the [NONCE$_2$] is the \underline{\hspace{0.8cm}}'' (no nonce in context)
\item \textbf{no\_preamble}: ``the [NONCE$_2$] is the \underline{\hspace{0.8cm}}'' (no context at all)
\end{itemize}

ME-consistent = $P(\text{NONCE}_2) > P(\text{FAM})$ at the completion position. 8 items $\times$ 10 embedding initialisation seeds per model. Nonce tokens use Strategy B embedding initialisation: each nonce is anchored to a random rare-noun embedding with 10\% Gaussian noise. This is the most conservative available strategy---centroid-based approaches produce artificially clustered nonce embeddings that inflate similarity priming, while purely random initialisation places nonces far from the noun region, making probability comparisons uninformative. Strategy B produces nonce embeddings that are plausibly noun-like but not systematically similar to each other.

\subsubsection{Dose--Response Track (H3)}

We vary the number of labelling repetitions (0, 1, 2, 3) to test whether the priming advantage---the log-probability difference between the labelled and unlabelled noun---increases monotonically with dose. Evaluated on 5 noun pairs $\times$ 5 model seeds per cell (25 item--seed pairs per cell).

\subsubsection{Scoring}

All scoring uses log-probability of target tokens at the completion position, extracted via teacher-forced next-token prediction. All target tokens are verified to be single BPE tokens in the shared vocabulary.

\subsection{Pilot Findings (Exploratory)}
\label{sec:pilot}

Three findings from the single-model pilot motivate the confirmatory design:

\paragraph{Pilot Finding 1: BabyBERTa cross-sentence insensitivity.} BabyBERTa's $\langle$mask$\rangle$ predictions are invariant to prior discourse context ($< 10^{-6}$ difference across all conditions). This is a consequence of the training regime---individual sentences with masked token prediction---not a limitation of masked LMs as an architecture class; masked LMs trained on concatenated multi-sentence inputs might show different sensitivity. ME evaluation via multi-sentence suppression design is therefore impossible for this specific checkpoint. We report this as a methodological finding.

\paragraph{Pilot Finding 2: AR repetition priming.} On the pilot AR model (6L-256d, PPL 36.4), all 6/6 familiar--familiar pairs show anti-ME: labelling FAM$_1$ increases $P(\text{FAM}_1)$ and decreases $P(\text{FAM}_2)$. Priming ranges from +1.3 to +2.9 nats. This is consistent with CDS statistics---parents repeatedly name objects (``that's a ball\,.\,see the ball\,.\,get the ball'').

\paragraph{Pilot Finding 3: Nonce ME is embedding similarity.} The context-dependence diagnostic reveals that nonce\_only (7/8 ME-consistent) exceeds full\_context (2/8). Introducing any nonce in context primes nonce-region tokens; introducing a familiar noun suppresses them via frequency advantage. The disambiguation context contributes nothing.

\subsection{Pre-Registered Hypotheses}
\label{sec:hypotheses}

Based on the pilot, we pre-registered four hypotheses:

\noindent\textbf{H1 (confirmatory):} Anti-ME rate $> 50\%$ in all 9 cells of the training grid. \emph{Test:} One-tailed sign test per cell, $\alpha = 0.05$. \emph{Criterion:} All 9 cells must reach significance.

\noindent\textbf{H2 (confirmatory):} nonce\_only $\geq$ full\_context in all 9 cells. \emph{Test:} Wilcoxon signed-rank test (5 paired seeds), $\alpha = 0.05$. \emph{Criterion:} All 9 cells.

\noindent\textbf{H3 (confirmatory):} $>$50\% of item--seed pairs show strictly monotonic dose--response in all 9 cells. \emph{Test:} Kendall's $\tau$ for trend, bootstrap 95\% CI on slope, $\alpha = 0.05$. \emph{Criterion:} All 9 cells $>$50\% monotonic.

\noindent\textbf{H4 (confirmatory):} Conjunction of H1: no cell shows ME-consistent suppression.

Each hypothesis is treated as a separate pre-registered question. No family-wise error rate correction is applied across hypotheses, as each addresses a distinct theoretical prediction with a distinct control comparison.

\section{Results}

We report confirmatory results followed by exploratory analyses. All confirmatory tests use the pre-registered statistical procedures and significance thresholds.

In summary: anti-ME repetition priming is significant in all 9 cells (H1 confirmed); the context-dependence diagnostic replicates in all 9 cells (H2 confirmed); dose--response priming shows significant positive trends in all 9 cells, with strict monotonicity in 8/9 (H3 partially confirmed); and no cell shows ME-consistent suppression (H4 confirmed). Priming attenuates with improved language modelling ($\rho = -0.533$) but never crosses zero.

\subsection{Training Grid}

All 45 models converged without exclusions. Table~\ref{tab:ppl} reports perplexity on a held-out AO-CHILDES test set.

\begin{table}[h]
\centering
\caption{Perplexity (mean $\pm$ SD across 5 seeds).}
\label{tab:ppl}
\small
\begin{tabular}{lccc}
\toprule
 & 5 epochs & 10 epochs & 20 epochs \\
\midrule
Small (2.9M) & 57.8 $\pm$ 0.1 & 48.5 $\pm$ 0.1 & 42.0 $\pm$ 0.1 \\
Medium (8.9M) & 46.3 $\pm$ 0.1 & 36.6 $\pm$ 0.1 & 28.9 $\pm$ 0.0 \\
Large (33.5M) & 39.3 $\pm$ 0.1 & 27.0 $\pm$ 0.1 & 15.2 $\pm$ 0.1 \\
\bottomrule
\end{tabular}
\end{table}

PPL spans a 3.8$\times$ range (15.2 to 57.8). Seed variance is negligible (all SDs $\leq$ 0.1).

\subsection{H1: Anti-ME at All Scales (Confirmed)}

Anti-ME rates range from 85\% to 100\%, with all sign tests significant at $p < 2.5 \times 10^{-13}$ (Table~\ref{tab:h1}). Not a single cell shows majority ME-consistent suppression.

\begin{table}[h]
\centering
\caption{H1 results. Anti-ME rate = proportion of item--seed pairs where labelling FAM$_1$ increases $P(\text{FAM}_1)$. Priming in nats (natural log units; negative = anti-ME direction).}
\label{tab:h1}
\small
\begin{tabular}{lccc}
\toprule
Cell & Anti-ME & Priming (nats) & Sign test $p$ \\
\midrule
Small, 5\,ep & 100/100 & $-1.14$ & $7.89 \times 10^{-31}$ \\
Small, 10\,ep & 100/100 & $-1.47$ & $7.89 \times 10^{-31}$ \\
Small, 20\,ep & 99/100 & $-1.49$ & $7.97 \times 10^{-29}$ \\
Medium, 5\,ep & 100/100 & $-1.46$ & $7.89 \times 10^{-31}$ \\
Medium, 10\,ep & 100/100 & $-1.43$ & $7.89 \times 10^{-31}$ \\
Medium, 20\,ep & 98/100 & $-0.81$ & $3.98 \times 10^{-27}$ \\
Large, 5\,ep & 100/100 & $-1.20$ & $7.89 \times 10^{-31}$ \\
Large, 10\,ep & 98/100 & $-1.06$ & $3.98 \times 10^{-27}$ \\
Large, 20\,ep & 85/100 & $-0.58$ & $2.41 \times 10^{-13}$ \\
\bottomrule
\end{tabular}
\end{table}

Priming attenuates with improved language modelling but never crosses zero. \textbf{H1 confirmed.}

\begin{figure}[t]
\centering
\includegraphics[width=0.65\columnwidth]{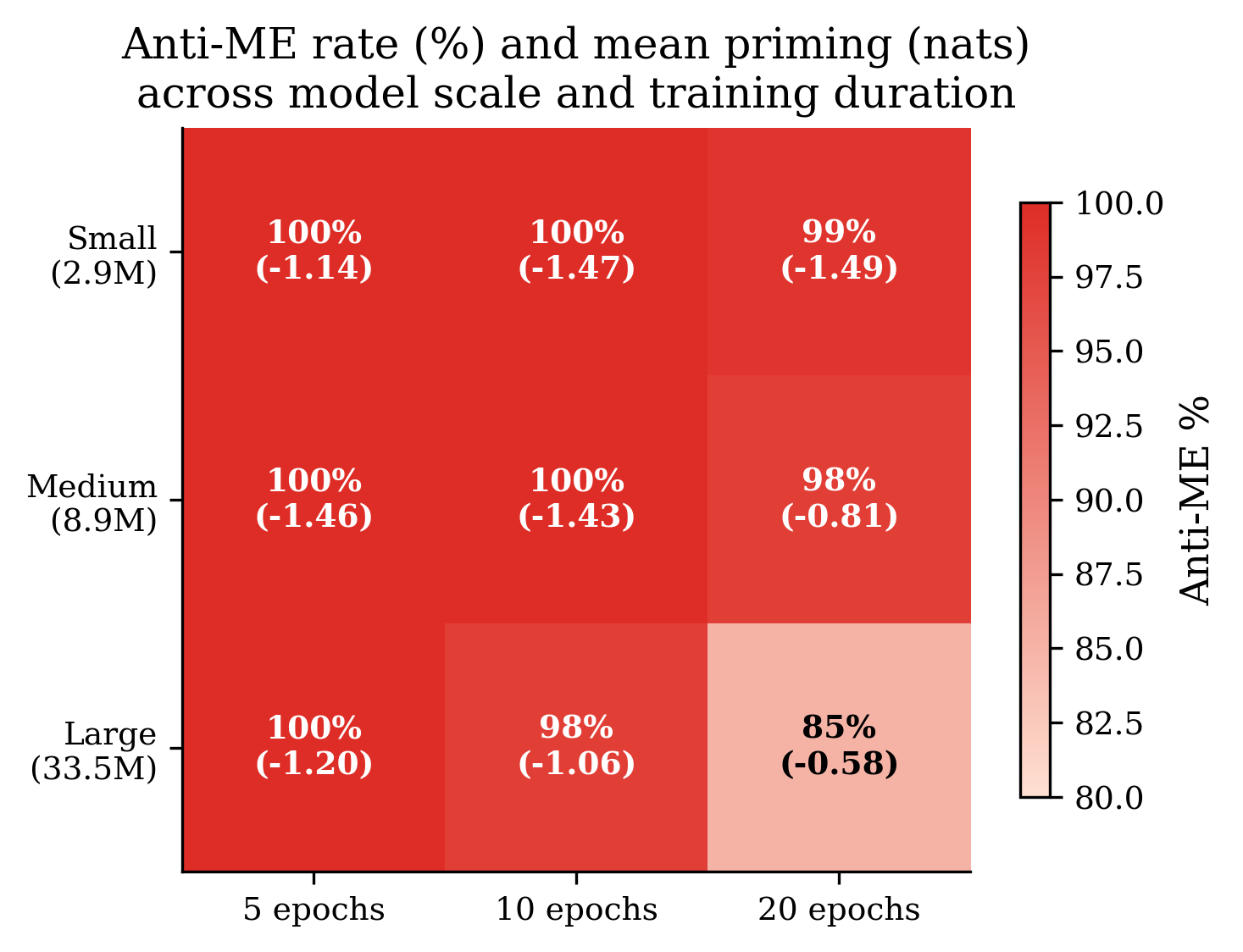}
\caption{Anti-ME rate (\%) and mean priming (nats) across model scale and training duration. All cells show strong anti-ME; priming magnitude decreases toward the bottom-right (larger, more-trained models).}
\label{fig:heatmap}
\end{figure}

\subsection{H2: Context-Dependence Diagnostic Replicates (Confirmed)}

nonce\_only exceeds full\_context in all 9 cells (Table~\ref{tab:h2}). All Wilcoxon tests reach $p = 0.0312$ (the minimum achievable for 5 paired observations all in the same direction).

Aggregated across all 45 models, the five-condition ordering is: nonce\_only (6.0/8) $>$ no\_preamble (3.1/8) $>$ full\_context (2.7/8) $>$ swap\_context (2.3/8) $>$ fam\_only (0.7/8). This gradient reflects embedding-region priming: nonce context boosts nonce-region predictions; familiar-noun context suppresses them. The disambiguation preamble adds nothing.

\begin{table}[h]
\centering
\caption{H2 results. Mean ME-consistent items out of 8.}
\label{tab:h2}
\small
\begin{tabular}{lccc}
\toprule
Cell & nonce\_only & full\_context & fam\_only \\
\midrule
Small, 5\,ep & 5.0 & 2.2 & ${\sim}$0 \\
Small, 10\,ep & 5.7 & 2.7 & ${\sim}$0 \\
Small, 20\,ep & 5.3 & 2.8 & ${\sim}$0 \\
Medium, 5\,ep & 6.7 & 2.4 & ${\sim}$0 \\
Medium, 10\,ep & 6.9 & 3.3 & ${\sim}$0 \\
Medium, 20\,ep & 6.4 & 3.2 & ${\sim}$0 \\
Large, 5\,ep & 6.0 & 2.0 & ${\sim}$0 \\
Large, 10\,ep & 6.2 & 2.9 & ${\sim}$0 \\
Large, 20\,ep & 5.7 & 2.6 & ${\sim}$0 \\
\bottomrule
\end{tabular}
\end{table}

\textbf{H2 confirmed.}

\begin{figure}[t]
\centering
\includegraphics[width=0.75\columnwidth]{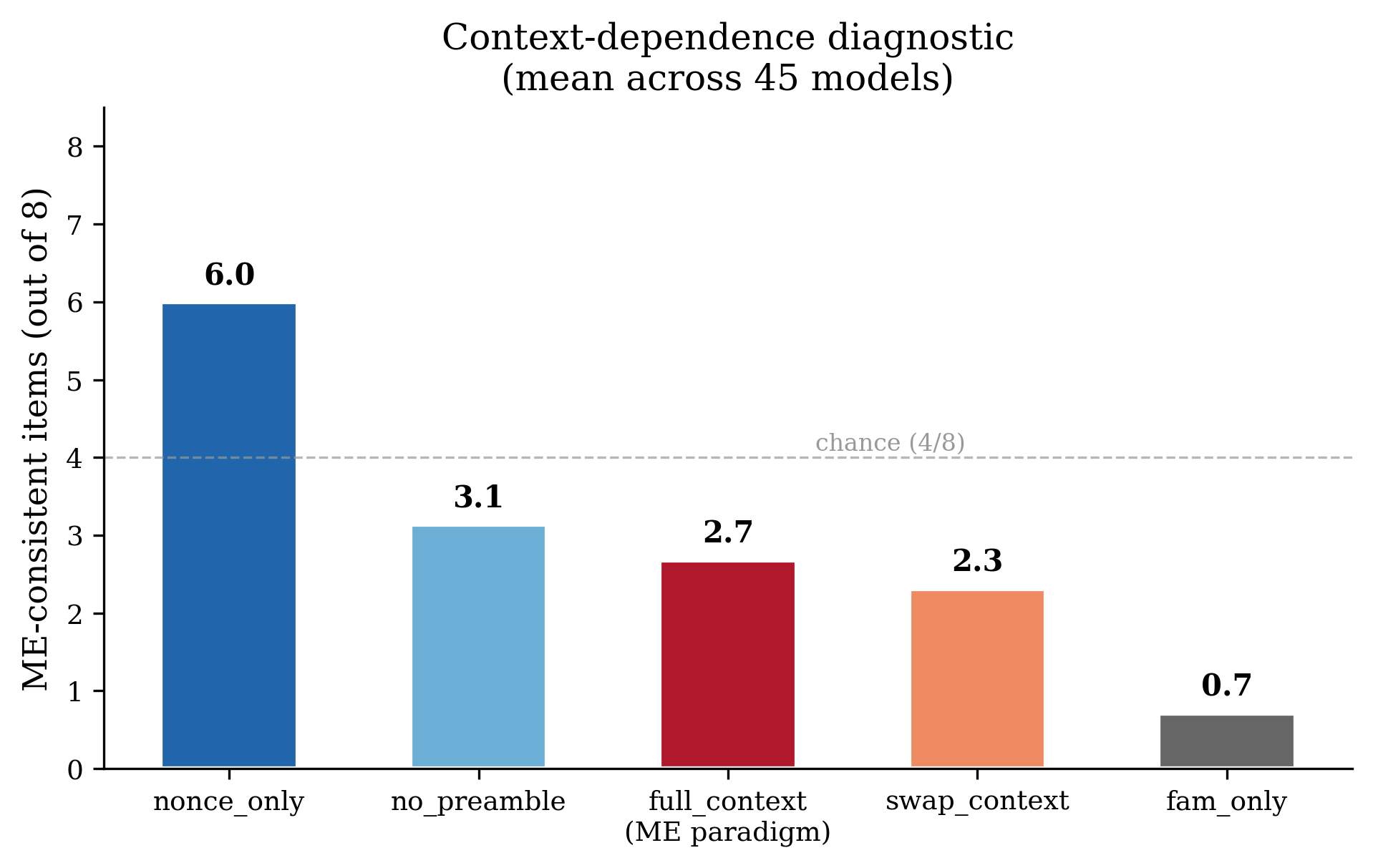}
\caption{Context-dependence diagnostic (mean across 45 models). The nonce\_only condition produces the highest apparent ``ME,'' despite containing no familiar noun and no ME paradigm. The full ME design (full\_context) scores below chance. Dashed line: chance (4/8).}
\label{fig:diagnostic}
\end{figure}

\subsection{H3: Dose--Response (Partially Confirmed)}

8/9 cells exceed the 50\% monotonic threshold. The exception is Large at 20 epochs (48\%, 12/25 pairs). All Kendall's $\tau$ are positive and significant (all $p < 0.002$); all bootstrap 95\% CIs on slope exclude zero (Table~\ref{tab:h3}).

\begin{table}[h]
\centering
\caption{H3 results.}
\label{tab:h3}
\small
\begin{tabular}{lcccc}
\toprule
Cell & Monotonic & Slope & Kendall $\tau$ & $p$ \\
\midrule
Small, 5\,ep & 25/25 & +0.576 & +0.391 & $2.07 \times 10^{-7}$ \\
Small, 10\,ep & 25/25 & +0.847 & +0.480 & $1.98 \times 10^{-10}$ \\
Small, 20\,ep & 22/25 & +0.965 & +0.421 & $2.39 \times 10^{-8}$ \\
Medium, 5\,ep & 25/25 & +1.080 & +0.497 & $4.19 \times 10^{-11}$ \\
Medium, 10\,ep & 24/25 & +1.138 & +0.451 & $2.13 \times 10^{-9}$ \\
Medium, 20\,ep & 15/25 & +0.896 & +0.241 & $1.36 \times 10^{-3}$ \\
Large, 5\,ep & 20/25 & +1.159 & +0.477 & $2.51 \times 10^{-10}$ \\
Large, 10\,ep & 15/25 & +0.976 & +0.391 & $2.14 \times 10^{-7}$ \\
Large, 20\,ep & 12/25 & +1.067 & +0.317 & $2.66 \times 10^{-5}$ \\
\bottomrule
\end{tabular}
\end{table}

Priming reliably increases with dose at all scales (all slopes positive, all trends significant), but strict item-level monotonicity weakens in the best language models. \textbf{H3 partially confirmed.}

\begin{figure}[t]
\centering
\includegraphics[width=\columnwidth]{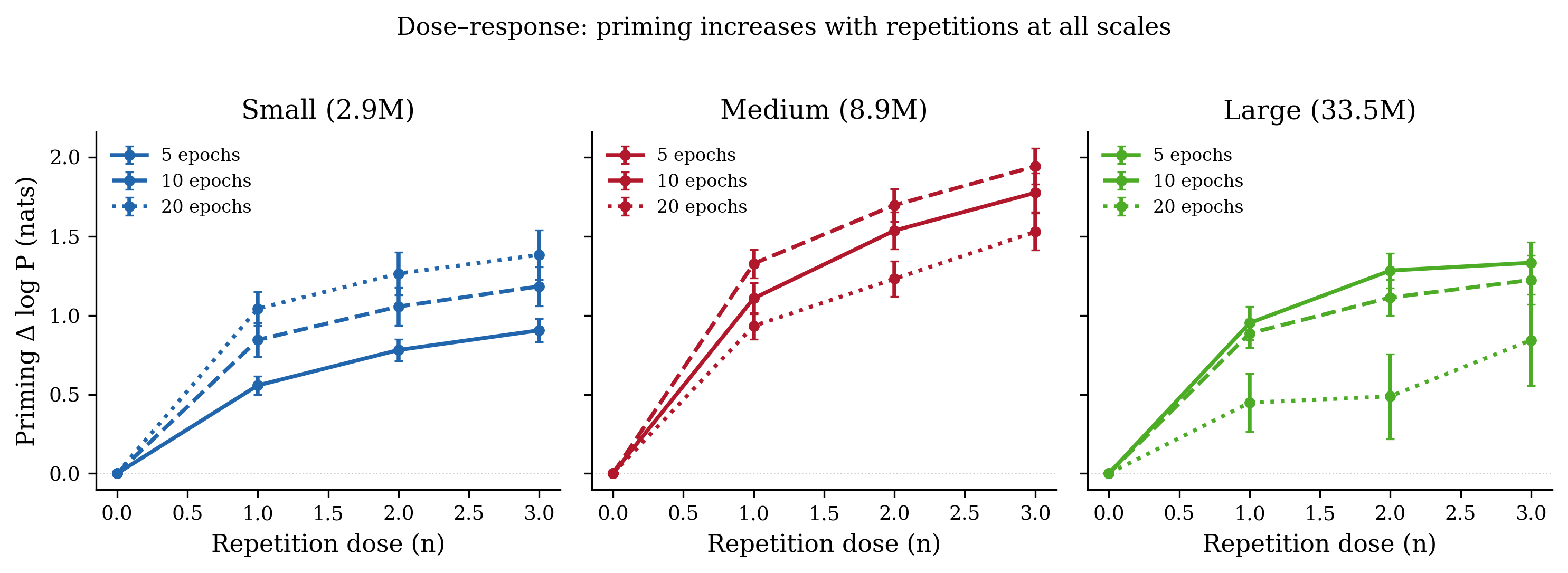}
\caption{Dose--response curves by model size. Priming advantage (labelled minus unlabelled log-probability) increases with repetition dose at all scales. Error bars: SEM across item--seed pairs. Concavity in the 20-epoch curves (dotted) reflects declining marginal returns on repetition.}
\label{fig:dose}
\end{figure}

\subsection{H4: No ME at Any Scale (Confirmed)}

Conjunction of H1. No cell shows ME-consistent suppression. \textbf{H4 confirmed.}

\subsection{Exploratory: PPL--Priming Relationship}

\begin{figure}[t]
\centering
\includegraphics[width=0.8\columnwidth]{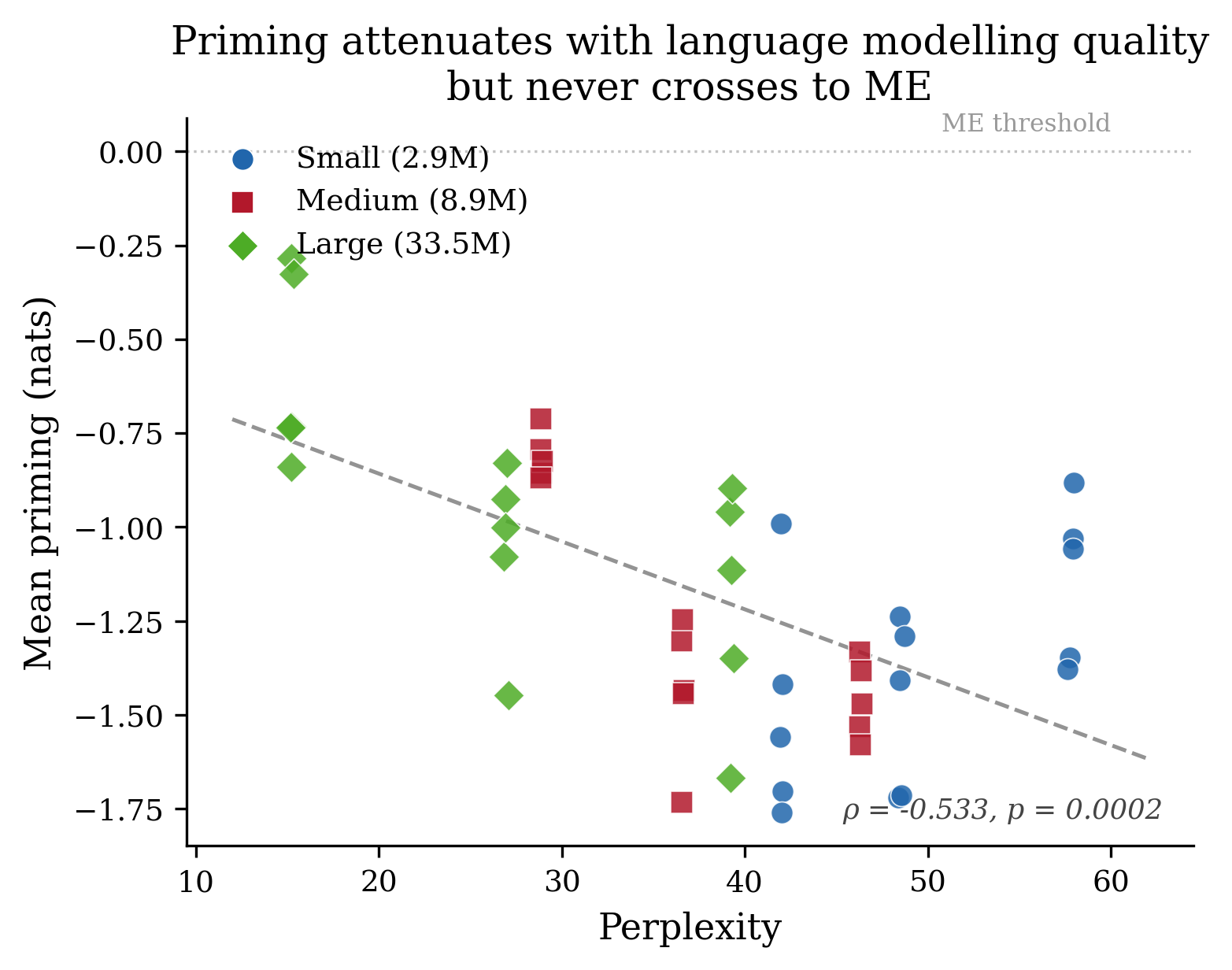}
\caption{PPL vs.\ mean priming across 45 models. Each point is one model; shapes indicate size class. Dashed line: linear regression. Dotted line at 0: ME threshold. Priming attenuates with lower perplexity but remains strictly negative (anti-ME).}
\label{fig:scatter}
\end{figure}

Across the 45 models, perplexity and priming magnitude are correlated (Spearman $\rho = -0.533$, $p = 0.0002$): better language models show weaker repetition priming. Decomposing this, model size contributes independently ($\rho = +0.434$, $p = 0.003$) as does training duration ($\rho = +0.335$, $p = 0.024$). The relationship is approximately log-linear ($R^2 = 0.44$, vs.\ linear $R^2 = 0.36$), with the linear intercept predicting $-0.50$ nats of anti-ME priming even at PPL $= 1$. Under the log-linear fit, priming would reach zero only at PPL $\approx 5.7$. For context, our best models achieve PPL 15.2 on held-out CDS after 20 epochs of training; reaching PPL 5.7 would require a model that predicts CDS tokens with substantially less uncertainty than our most overtrained configuration, a regime more consistent with corpus memorisation than with linguistic generalisation. The trajectory is one of attenuation toward a non-zero asymptote, not convergence toward ME (Figure~\ref{fig:scatter}).

\subsection{Exploratory: Per-Item Analysis}

Items vary in priming magnitude. The strongest priming pair is cup/book (mean $-2.10$ nats, 98\% anti-ME across all 45 models). The weakest are dog/cat ($-0.42$ nats, 96\% anti-ME) and cat/dog ($-0.48$ nats, 93\% anti-ME). Both weak-priming pairs have natural synonyms in AO-CHILDES (dog/puppy, cat/kitty), potentially diluting the repetition signal. This parallel to bilingual attenuation of ME \citep{byersheinlein2009monolingual} is suggestive but requires targeted follow-up.

Per-noun corpus frequency does not predict per-noun priming magnitude (Spearman $\rho = 0.07$, $p = 0.86$, $n = 9$), nor does per-noun repetition rate ($\rho = -0.15$, $p = 0.70$). This suggests that while CDS repetition structure explains the \emph{direction} of the effect (priming rather than suppression), per-item variation in magnitude is driven by other factors---likely properties of the unlabelled competitor noun or pair-specific co-occurrence patterns rather than the labelled noun's frequency alone.

\section{Discussion}

\subsection{Repetition Priming Without Mutual Exclusivity}

The central finding is a dissociation: autoregressive language models trained on child-directed speech develop robust repetition priming but do not develop mutual exclusivity at any scale tested. Across a 3.8$\times$ perplexity range (15.2 to 57.8), a 12$\times$ parameter range (2.9M to 33.5M), and training durations from 5 to 20 epochs, every model shows anti-ME behaviour.

The PPL--priming relationship ($\rho = -0.533$) is informative about trajectory. Better language models attenuate their repetition priming---the Large, 20-epoch models show roughly half the priming magnitude of the Small, 5-epoch models---but the trajectory converges toward a non-zero asymptote rather than crossing to suppression. The log-linear fit extrapolates to zero priming only at PPL $\approx 5.7$, well below the perplexity achievable by models that generalise; the linear intercept predicts $-0.50$ nats even at PPL $= 1$. This addresses a natural objection: that ME might emerge in a sufficiently capable model. While we cannot rule out ME at scales far beyond those tested here, the fitted attenuation curve provides no indication of an approaching sign change. The data are more consistent with a stable equilibrium between learned repetition statistics and improved discourse modelling than with a trajectory toward ME.

\subsection{The H3 Nuance: Monotonicity Weakens in the Best Models}

The partial disconfirmation of H3 deserves careful interpretation. All cells show a significant positive dose--response trend, but strict item-level monotonicity weakens in the best models (48\% for Large at 20 epochs). We consider three possible interpretations.

First, this could reflect noise---but the effect is systematic, appearing in all three 20-epoch cells and correlating with perplexity, which argues against random variation.

Second, one might ask whether this is an early sign of ME-like behaviour. We think not, for two reasons: the aggregate trend remains positive (more repetitions still produce more priming on average), and the non-monotonicity appears as local reversals at high dose rather than a sign change at low dose. A dose-step analysis confirms this: the first repetition (dose $0 \to 1$) increases priming in 60--100\% of item--seed pairs across all cells, including 60\% for Large at 20 epochs. Non-monotonicity concentrates at steps $1 \to 2$ and $2 \to 3$, where 20-epoch models show 60--80\% positive increments compared to 96--100\% in 5-epoch models. ME would predict suppression at dose 1, which is never the majority pattern in any cell.

Third, and most consistent with the data, the weakening of strict monotonicity likely reflects improved discourse modelling. AO-CHILDES contains natural topic shifts: caregivers do not repeat a single noun indefinitely. A model that has learned this distributional regularity should assign declining marginal probability to each additional repetition---producing a concave dose--response curve that violates strict monotonicity while preserving the positive trend. This interpretation is supported by the observation that the effect is strongest in our best language models (PPL 15.2), which have the most capacity to learn discourse-level statistics.

\subsection{Grounding as the Missing Ingredient}

Our findings converge with the grounded models literature. \citet{nortje2024mutual} demonstrated ME in visually grounded speech models. \citet{oneata2025mutual} showed weaker ME in bilingual VGS models, paralleling human data. \citet{thai2025mebench} found weak ME bias even in state-of-the-art VLMs. The pattern across these results and ours is consistent: ME appears to require referential grounding. When a model has referents to map labels to, ME can emerge; when it operates over text alone, the mechanisms available---embedding similarity and discourse-frequency effects---do not produce lexical exclusivity. This is consistent with the Bayesian framework's requirement for a referent set, but we note that our evidence is correlational across studies rather than a direct manipulation of grounding within a single architecture.

This connects to the Bayesian word-learning framework \citep{xu2007word}, which formalises ME as a consequence of probabilistic inference over word--referent mappings. The key computational ingredient is the referent set. \citet{lewis2020role} showed that ME in children is graded, probabilistic, and modulated by experience---precisely what a Bayesian framework predicts. Text-only models lack the referent set entirely. Taken together, the evidence suggests that distributional statistics in CDS favour repetition-based discourse tracking, while exclusivity emerges only when learners---whether human or computational---represent competing referents that must be disambiguated.

\subsection{The Context-Dependence Diagnostic}

Beyond the ME-specific findings, the context-dependence diagnostic introduced here has broader utility. The logic is simple: if an apparent pragmatic effect survives the removal of the pragmatically relevant context, it is not pragmatic. In our case, stripping the disambiguation preamble while retaining nonce context actually \emph{increased} the apparent ME signal---revealing embedding similarity as the true driver. The same ablation logic applies to any prompt-based evaluation claiming pragmatic competence in LMs, whether for scalar implicature, presupposition accommodation, or other phenomena where low-level token effects might masquerade as inference.

\subsection{Scope of the Claim}

An important clarification: we do not claim that distributional learning is fundamentally inadequate for referential pragmatics, nor that ME requires innate linguistic machinery. Our claim is narrower---that the modality of input constrains what referential mechanisms can be learned. ME involves word--referent mapping, and mapping requires referents. Under child-scale data budgets and text-only input, the distributional signal in CDS favours repetition over exclusivity. Whether different architectures, training objectives, or much larger scales could overcome this is an open empirical question; our data provide no evidence that they would, but neither do they rule it out. This framing is consistent with the BabyLM programme's multimodal track \citep{warstadt2023findings,charpentier2025findings}, which recognises that some aspects of language competence may require multimodal input.

\subsection{Limitations}

Several constraints bound our conclusions. We tested a single corpus (AO-CHILDES, ${\sim}$4M tokens); the full BabyLM 10M corpus includes books, subtitles, and other genres whose distributional statistics may differ from CDS in ways that matter for referential behaviour. Our largest model has 33.5M parameters---large for the BabyLM regime but small compared to modern LMs. The monotonic attenuation trajectory gives no hint of an approaching phase transition, but we cannot rule out qualitative change at scales orders of magnitude beyond ours. On the methodological side, nonce embedding initialisation (Strategy B) influences the magnitude of nonce-related effects, though not their direction. Finally, text-only CDS transcripts are not entirely devoid of referential structure---speaker turns and deictic expressions carry implicit information about the discourse situation---and our BabyBERTa finding is specific to a checkpoint trained on isolated sentences; masked LMs with multi-sentence training might behave differently.

\section{Conclusion}

Child-scale language models trained on CDS develop repetition-based reference tracking rather than mutual exclusivity. The dissociation is robust across a 12$\times$ parameter range and 3.8$\times$ perplexity range, with priming attenuating but not reversing as model quality improves. The pattern is consistent with the grounded cognition framework: ME involves word--referent mapping, and models without referents lack the computational substrate for it. Our context-dependence diagnostic provides a reusable tool for distinguishing genuine pragmatic inference from embedding-level artefacts in LM evaluations. The referential gap in BabyLM evaluation is not merely methodological---it reflects a structural limitation of what text-only distributional learning can achieve at these scales, and motivates multimodal approaches for referential phenomena.

\bibliographystyle{plainnat}
\bibliography{references}

\end{document}